\pdfoutput=1
\documentclass[11pt]{article}

\usepackage[]{acl}

\usepackage{times}
\usepackage{latexsym}
\usepackage[T1]{fontenc}
\usepackage[utf8]{inputenc}
\usepackage{microtype}

\usepackage{xcolor,colortbl}
\usepackage{hyperref}
\usepackage{graphicx}
\usepackage{adjustbox}
\usepackage{amssymb}
\usepackage{pifont}
\usepackage{multirow}
\usepackage{amsmath}
\usepackage{bbm}
\usepackage{float}
\usepackage{caption}
\usepackage{subcaption}
\usepackage{pifont}
\newcommand{\cmark}{\ding{51}}%
\newcommand{\xmark}{\ding{55}}%

\usepackage{booktabs}

\definecolor{cinnabar}{rgb}{0.89, 0.26, 0.2}
\definecolor{citrine}{rgb}{0.89, 0.82, 0.04}
\definecolor{cobalt}{rgb}{0.0, 0.28, 0.67}
\definecolor{darkorchid}{rgb}{0.6, 0.2, 0.8}
\definecolor{darkpastelgreen}{rgb}{0.01, 0.75, 0.24}

\title{
Analyzing and Evaluating Faithfulness in Dialogue Summarization
}

\author{
Bin Wang\textsuperscript{\rm \dag}, 
Chen Zhang\textsuperscript{\rm \dag}, 
Yan Zhang\textsuperscript{\rm \dag}, 
Yiming Chen\textsuperscript{\rm \dag}, 
Haizhou Li\textsuperscript{\rm $\natural$, \rm \dag, \rm $\S$}
\\
\textsuperscript{\rm \dag}National University of Singapore, Singapore
\\
\textsuperscript{\rm $\natural$}The Chinese University of Hong Kong, Shenzhen, China
\\
\textsuperscript{\rm $\S$}Kriston AI, China
\\
\texttt{bwang28c@gmail.com}
}

\begin{document}
\maketitle
\begin{abstract}
    Dialogue summarization is abstractive in nature, making it suffer from factual errors. The factual correctness of summaries has the highest priority before practical applications. Many efforts have been made to improve faithfulness in text summarization. However, there is a lack of systematic study on dialogue summarization systems. In this work, we first perform the fine-grained human analysis on the faithfulness of dialogue summaries and observe that over $35\%$ of generated summaries are faithfully inconsistent respective the source dialogues. Furthermore, we present a new model-level faithfulness evaluation method. It examines generation models with multi-choice questions created by rule-based transformations. 
    Experimental results show that our evaluation schema is a strong proxy for the factual correctness of summarization models.
    The human-annotated faithfulness samples and the evaluation toolkit are released to facilitate future research toward faithful dialogue summarization. Code available: \url{https://github.com/BinWang28/FacEval}. \footnote{Accepted in EMNLP 2022}

\end{abstract}

\section{Introduction}

    Text summarization aims to condense a document into a short paragraph or a single sentence while conveying the core information \cite{el2021automatic}. It can be either extractive or abstractive. Extractive summarization methods identify salient sequence spans from the source document and pasting them together \cite{dorr2003hedge,kobayashi2015summarization,zhong2020extractive}. Abstractive summarization methods generate completely new summary in a coherent manner \cite{paulus2017deep,lewis2019bart,zhang2020pegasus,liu2021topic,wang2022focused}. Previous work discovered that abstractive summarization suffers from unfaithful outputs, limiting its applicability in real-world scenarios \cite{kryscinski2019evaluating,falke2019ranking,zhu2020enhancing,ladhak2021faithful}.

    \begin{table}[t]
      \centering
      \begin{adjustbox}{width=0.48\textwidth,center}
        \begin{tabular}{ | l | }
            \hline
            \emph{\textbf{Dialogue}}: \\ \hline
            \textcolor{darkorchid}{\emph{Freddie}}: Nanna, are you coming to visit us soon? 
            \textcolor{darkorchid}{\emph{Winnie}}: \\
             Oh darling, Nanna has broken her leg, you'll have to visit \\
             me instead. \textcolor{darkorchid}{\emph{Freddie}}: I forgott. Well come soon. \textcolor{darkorchid}{\emph{Winnie}}: \\
              Good, ask Mummy and Daddy and they will come when \\
             they can. \textcolor{darkorchid}{\emph{Freddie}}: Yes love you. Leg better soon? \textcolor{darkorchid}{\emph{Winnie}}: \\
              Yes, quite soon. Tell mummy to ring me. Bye darling xxxxx \\ \hline
            \emph{\textbf{Summaries}}: \\\hline
            \textcolor{teal}{\emph{Human}}: Winnie has broken her leg and will not visit \\ any time soon. Freddie will ask mummy to call Winnie up. \cmark  \\ \hline
            \textcolor{teal}{\emph{BART}}: Nanna has broken her leg, so Freddie will have to visit \\ her instead. Nanna will get better soon. \cmark  \\ \hline
            \textcolor{teal}{\emph{MV-BART}}: Nanna has broken her leg and Freddie will have to \\visit Winnie instead. Mummy and Daddy will come to visit \\ \underline{them} soon.  \xmark \\ \hline
            \textcolor{teal}{\emph{Coref-BART}}: \underline{Freddie wants to visit Winnie}, but Nanna has \\ broken her leg, so he will have to visit her instead. Mummy \\ and Daddy will come when they can.  \xmark \\ \hline
            \textcolor{teal}{\emph{CondigSum-BART}}: \underline{Winnie's Nanna} has broken her leg and\\ Freddie will have to visit her instead.  \xmark \\ \hline
        \end{tabular}
      \end{adjustbox}
      \caption{A real example from SAMSum dataset. Span of factual errors are marked with \underline{underline}.}
      \label{tab:example}
    \end{table}
    
    As an essential way of exchanging information, conversations usually involve multiple participants, informal language usage, repetition, and negations \cite{sacks1978simplest,chen2020multi}.
    Therefore, dialogue summarization is vulnerable to factual issues due to its abstractive nature. Table~\ref{tab:example} gives an example of factually incorrect dialogue summaries. The problem of factual correctness is broadly studied for text summarization in news and article domains \cite{nallapati2016abstractive,narayan2018don}.
    The progress is primarily because of the availability of factually annotated data at both summary and token levels \cite{kryscinski2019evaluating,wang2020asking,pagnoni2021understanding,cao2022hallucinated}. Many studies are proposed to evaluate and reduce factual errors in the generated summaries.
    However, due to the interactive nature of dialogues, we cannot simply transfer these methods to dialogue summarization.

    In this work, we first categorize the most frequently occurred factual errors for dialogue summarization into $6$ types. Then, we collect fine-grained factual annotations for human reference and the output of $4$ recent dialogue summarization systems ($\S$\ref{sec:analysis}). At least two annotators are involved, and a verification process is incorporated to ensure the annotation quality. As a result, our study on human-annotated data suggests that over $35\%$ of the generated dialogue summaries contain at least one factual error. Similar observations have been made in the news summarization domain where $30\%$-$80\%$ of generated text are factually inconsistent \cite{cao2018faithful,pagnoni2021understanding}. More research attention should be made toward faithful dialogue summarization.
    
    The unavailability of faithful evaluation methods hinders the development of effective dialogue summarization models. In this work, we present a model-level evaluation schema, FacEval, targeting dialogue summarisation models' faithfulness ($\S$\ref{sec:faceval}). First, we synthesize a set of positive and negative summaries for each dialogue with back-translation or rule-based transformations. Then, a summarization model is asked to distinguish positive and negative summaries based on conditional generation probabilities. More correct judgements indicate the model is more factually competent.
    
    To compare the model-level performance of evaluation methods, we leverage two \emph{ad-hoc} training schema to synthesize a series of models with different capability ranks. Then, the evaluation methods are used to predict the ranking of trained models. Seven non-factual and factual evaluation methods have been examined, followed by a detailed discussion of their properties. The effectiveness of FacEval is also proven by showing a strong correlation with the factual correctness of summarization models.

    \begin{table*}[t]
        \centering
        \begin{subtable}[t]{0.85\textwidth}
            \begin{adjustbox}{width=1.0\textwidth,center}
            \begin{tabular}{ c | c }
                \multicolumn{1}{l}{\textbf{Speaker 1:} Fiona} & \multicolumn{1}{l}{\textbf{Speaker 2:} Jonathan} \\
                \hline
                \rowcolor{gray!18} \emph{What should I prepare 4 my dad's birthday?}            &  \emph{How old is he?} \\ 
                 \emph{Turning 50.}        & \emph{Wow, a round birthday, it must be sth big.} \\
                \rowcolor{gray!18}  \emph{I know, but I don't have any idea.}                    &  \emph{What does he like?} \\ 
                  \emph{He watches a lot of military movies.}       & \emph{Well, a movie ticket is probably not what you thought of.} \\
                \rowcolor{gray!18} \emph{No, not even close.}                                       &  \emph{U said he likes military... maybe paintball?} \\
                \emph{I don't know how my mum will react but I like it :D} & \\
                \hline
            \end{tabular}
            \end{adjustbox}
            \vspace{2mm}
        \end{subtable}
        \quad
        \begin{subtable}[t]{1.0\textwidth}
                \begin{adjustbox}{width=1.0\textwidth,center}
                    \begin{tabular}{r l}
                    \rowcolor{blue!10} \multirow{1}{*}{\textbf{Ref. Summary}:}
                                    &   \textbf{\underline{\textcolor{violet}{Fiona}}} doesn't know what \textbf{\underline{\textcolor{magenta}{she}}} should give to her dad as \textbf{\underline{\textcolor{teal}{a birthday gift}}}. \textbf{\underline{\textcolor{orange}{He likes}}} military. \textbf{\underline{\textcolor{olive}{Jonathan suggests a paintball match.}}} \vspace{1mm} \\
                    \multirow{1}{*}{\textbf{SubObjE}:}
                                    &   \textbf{\textcolor{violet}{Jonathan}} doesn't know what she should give to her dad as a birthday gift. 
                                       He likes military. Jonathan suggests a paintball match. \\
                    \multirow{1}{*}{\textbf{ProE}:}
                                    &   Fiona doesn't know what \textbf{\textcolor{magenta}{he}} should give to her dad as a birthday gift. He likes military. Jonathan suggests a paintball match. \\
                    \multirow{1}{*}{\textbf{NegE}:}   
                                    &   Fiona doesn't know what she should give to her dad as a birthday gift. 
                                       \textbf{\textcolor{orange}{He hates}} military. Jonathan suggests a paintball match. \\ 
                    \multirow{1}{*}{\textbf{ParE}:}
                                    &   Fiona doesn't know what she should give to her dad as \textbf{\textcolor{teal}{a Christmas gift}}. 
                                       He likes military. Jonathan suggests a paintball match. \\ 
                    \multirow{1}{*}{\textbf{HalE}:} 
                                    &   Fiona doesn't know what she should give to her dad as a birthday gift. 
                                      He likes military. \textbf{\textcolor{olive}{Jonathan invites Fiona to watch a military movie.}} \\
                    \end{tabular}
                \end{adjustbox}
        \end{subtable}
        \caption{An illustration of the taxonomy on factual error types.
        }
        \label{tab:error_example}
    \end{table*}

\section{Related Work}
    
    \subsection{Summarization Methods}
    
        Text summarization is one of the most important tasks in natural language generation (NLG). With the development of pre-trained language models, a lot progress has been made to abstractive text summarization \cite{see2017get,zhang2020pegasus,liu2022brio}, especially in news domain \cite{hermann2015teaching,narayan2018don}. With the availability of datasets \cite{carletta2005ami,gliwa2019samsum,zhu2021mediasum}, dialogue summarization research has attracted a lot of attention. For dialogue summarization, fine-tuning pre-trained generation models including T5 \cite{raffel2019exploring}, PEGASUS \cite{zhang2020pegasus} and BART \cite{lewis2019bart} are served as a strong baseline, where BART achieves the SOTA performance on ROUGE scores. Some recent works consider the dialogue properties for more advanced summarization models. \citet{chen2020multi} and \citet{liu2021topic} incorporate the conversational structures into the semantic encoding process of dialogue. Conversations involve lots of co-references. Therefore, \citet{liu2021coreference} proposes injecting co-reference information into the transformer layers by adapting attention maps or through graph convolutional networks~(GCN). We include the outputs of recent dialogue summarization models in our analysis.

    
    \subsection{Faithfulness Analysis}
    
        Previous works spot that the factual consistency problem is one key aspect of improving text summarization \cite{kryscinski2019evaluating,cao2020factual}. The analysis of factual errors in summaries is mainly performed in the news domain. \citet{kryscinski2019neural} and \citet{falke2019ranking} conducted the initial crowdsourcing of binary factual annotations and found that nearly 30\% of the generated summaries are factually inconsistent. Recent extensions focus on more fine-grained analysis \cite{cao2021cliff,pagnoni2021understanding} and also discovering factual evidences at entity level \cite{cao2022hallucinated} or span level \cite{huang2020have,maynez2020faithfulness,goyal2021annotating}. 
        
        Recently, CONFIT presented the first study on the faithfulness of dialogue summaries \cite{tang2021confit}. Similar to our work, they also define a taxonomy of factual errors and conduct fine-grained annotations. However, they focus on comparing reference summaries and generated summaries without referring to the whole dialogue. It is sub-optimal because the reference summary cannot fully represent the entire dialogue and also can be incorrect according to our analysis in Section~\ref{sec:analysis}. Besides, the missing and redundant information is categorized as factual errors, which we consider less proper. More recent advanced dialogue summarization models are also not included in their analysis.
        
        
    \subsection{Faithfulness Evaluation}

        The default evaluation metric for summarization, ROUGE, is based on n-gram overlaps between a generated summary and the corresponding references, rendering it less sensitive for capturing factual errors. Therefore, several new metrics are proposed to evaluate the faithfulness in the news domain \cite{kryscinski2019neural,fabbri2021summeval,tang2022understanding}. There are two major groups, one is based on natural language inference, and the other is based on question-answering. \citet{kryscinski2019evaluating} and \citet{goyal2020evaluating} propose to leverage entailment relationship. \citet{scialom2021questeval} and~\citet{wang2020asking} involves question generation, answer generation and answer-overlap as the factual consistency measure. \citet{zhao2021todsum} proposes to evaluate the faithfulness of task-oriented dialogue summarization by calculating the amount of overlapped dialogue states, which requires additional human annotations. 
    

    

    
    
    
    
    
    
    
    
    
    
    

    
    


\begin{figure*}[t]
 \centering
     \includegraphics[width=0.95\textwidth]{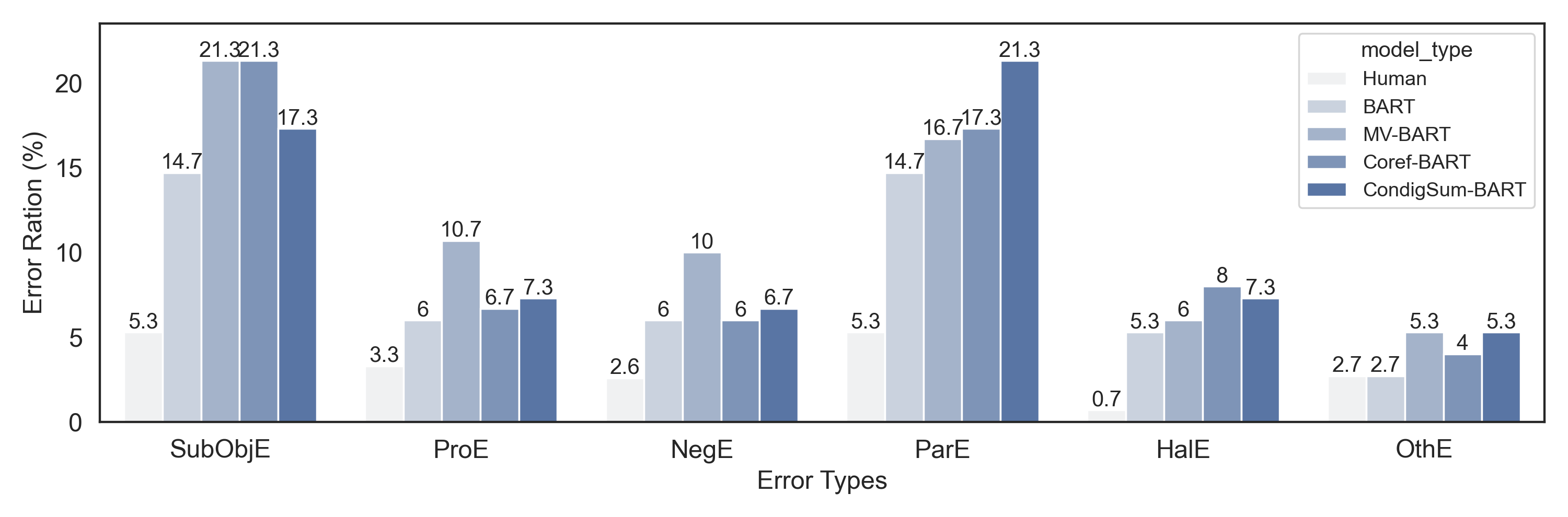}
\caption{The proportion of summaries with different types of factual errors. Note that one summary can contain multiple error types.}
\label{fig:error_ratio}
\end{figure*}

\begin{figure}[t]
 \centering
     \includegraphics[width=0.48\textwidth]{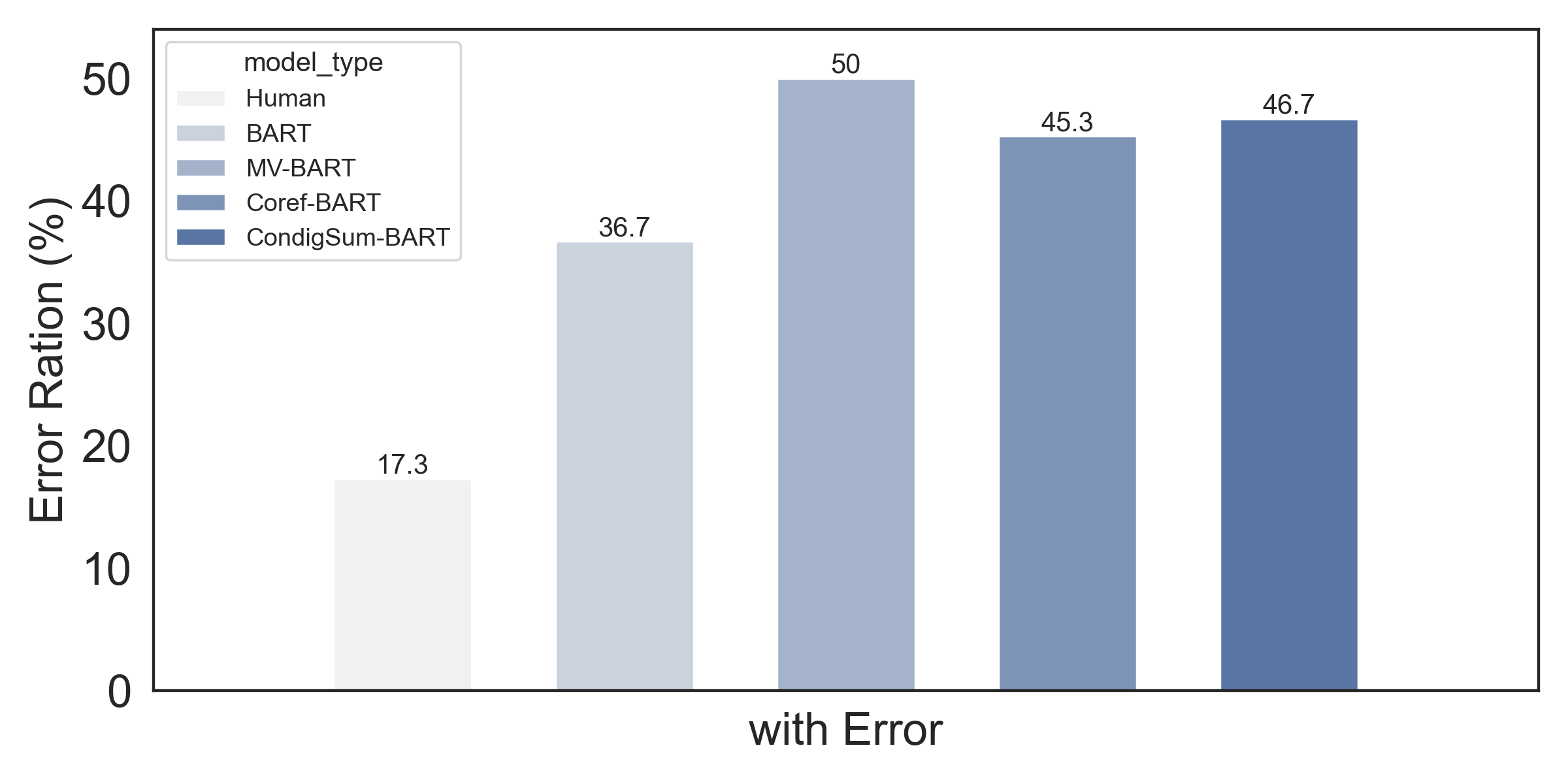}
\caption{The proportion of summaries with at least one factual error.}
\label{fig:all_error_ratio}
\end{figure}

\section{Fine-grained Faithfulness Analysis}
\label{sec:analysis}
    
    Previous studies of factuality analysis in summarization mainly focus on the news domain. The typology of factual errors for dialogues can be very different. Therefore, we first define a taxonomy of frequently occurred factual errors for dialogue summaries. A fine-grained analysis is then performed by measuring the factual consistency within dialogue summary pairs.
    
    
    
    
    

    \subsection{Taxonomy of Factual Errors}
    
        We collect the generated summaries using four SOTA dialogue summarization models on the popular dialogue summarization dataset, SAMSum \cite{gliwa2019samsum}. The selected models are BART \cite{lewis2019bart}, MV-BART \cite{chen2020multi}, Coref-BART \cite{liu2021coreference} and CondigSum-BART \cite{liu2021topic}. We define five most frequently occurred error types in dialogue summaries as below. An example for each error type is shown in Table~\ref{tab:error_example}.
        
        \noindent\textbf{Subject Object Error (SubObjE)}: The subject(s) or object(s) involved for an event is (partially) wrong. It includes substitution, addition and deletion of any related subject(s) or object(s).

        \noindent\textbf{Pronoun Error (ProE)}: Pronoun references are frequently occurred in dialogue summarization. This error includes wrong references and ambiguous ones that cannot be fully understandable relying on the summary.

        \noindent\textbf{Negation Error (NegE)}: Dialogues can contain confirmation utterances. This error means that the generated summary makes wrong conclusions when contradictory or unconfirmed events are presented in the dialogue.

        \noindent\textbf{Particulars Error (ParE)}: The summary presents related events, but some details are inaccurate or faulty. It can include incorrect information like date, time and location.

        \noindent\textbf{Hallucination Error (HalE)}: Generation models have the imaginary ability and can be triggered by certain prompt words in the dialogue. The hallucination error refers to the cases where the summary contains events not presented in the dialogue.

        \noindent\textbf{Other Error (OthE)}: It is used to classify factual errors that do not belong to any of the above types.
        
        Note that the above-mentioned error types are not exclusive to each other. That is, one summary may contain multiple error types.

    \subsection{Annotation Process}
        
        We random sample 150 dialogues from the test set of SAMSum. Five summaries are listed for each dialogue, including the human-written one and four model-generated summaries. 
        
        \citet{falke2019ranking} founds that it needs at least 12 annotations to reach an inter-annotator agreement of coefficient $k=0.75$, which can lead to high annotation costs and unreliable results with fewer annotators \cite{kryscinski2019evaluating}. Therefore, we perform a two-step verification process to ensure the annotation quality. First, each sample is annotated by two distinct annotators. If there is a disagreement about whether a summary contains factual errors, a third annotator is involved in making the final decision while considering inputs from the previous two annotators. As a result, we have collected 750 fine-grained faithfulness annotations from 30 participants.
        
        
        
        
        
        
        
        
        
        

    \subsection{Results and Analysis}
    
        The detailed annotation results are shown in Figure~\ref{fig:error_ratio}. There are several exciting findings: 
        1) the human annotations contain non-negligible factual errors at around 17\%; 
        2) 36\% to 50\% of generated summaries from dialogue summarization models contain at least one factual error;
        3) three advanced dialogue summarization models perform worse than their baseline on factual consistency.
        
        
        First, the popular SAMSum dataset \cite{gliwa2019samsum} associates each dialogue with one human-written reference summary. However, we found that 17\% of reference summaries have factual errors. Therefore, we encourage people to be aware of the issue, especially for evaluation. It is because the dialogue annotation process for SAMSum only involved one annotator per sample, and no further verification process was executed. We notice that the source of factual errors for human summaries is also different from machine-generated ones. Some factual errors in human-written summaries are caused by typos, which rarely occur in machine-generated summaries.
        
        For dialogue summarization models, we found that 35\%-50\% of generated summaries contain factual errors. The most frequent error types are SubObjE and ParE. Because dialogue often involves scattered information exchange with multiple speakers in multiple turns, it is very challenging to accurately locate \emph{who} and \emph{whom} in \emph{who-did-what-to-whom}. That is the leading cause of SubObjE. ParE is the second most frequent error type, indicating that the generated summaries express the same topic but do not accurately capture the details. OthE occurs less frequently. It shows that our taxonomy of factual errors can cover the most frequent error types for dialogue summarization.
      
        Surprisingly, we found that MV-BART, Coref-BART and CondigSum-BART perform even worse than the baseline model, with an increase of around 10\% overall factual error rate. They are accepted as more advanced summarization models and perform better on ROUGE scores. It indicates that enhancing topical information is not necessarily contributing much to factuality \cite{chen2020multi,liu2021topic}. Coref-BART aims to improve BART with co-reference information \cite{liu2021coreference}. However, our result shows it does not bring obvious benefits. In conclusion, we encourage the future development of summarization models to pay more attention to the factuality perspective, and a more diverse evaluation schema beyond ROUGE scores should be incorporated.
                
        

\begin{figure}[t]
 \centering
 \begin{subfigure}[b]{0.45\textwidth}
     \centering
     \includegraphics[width=\textwidth]{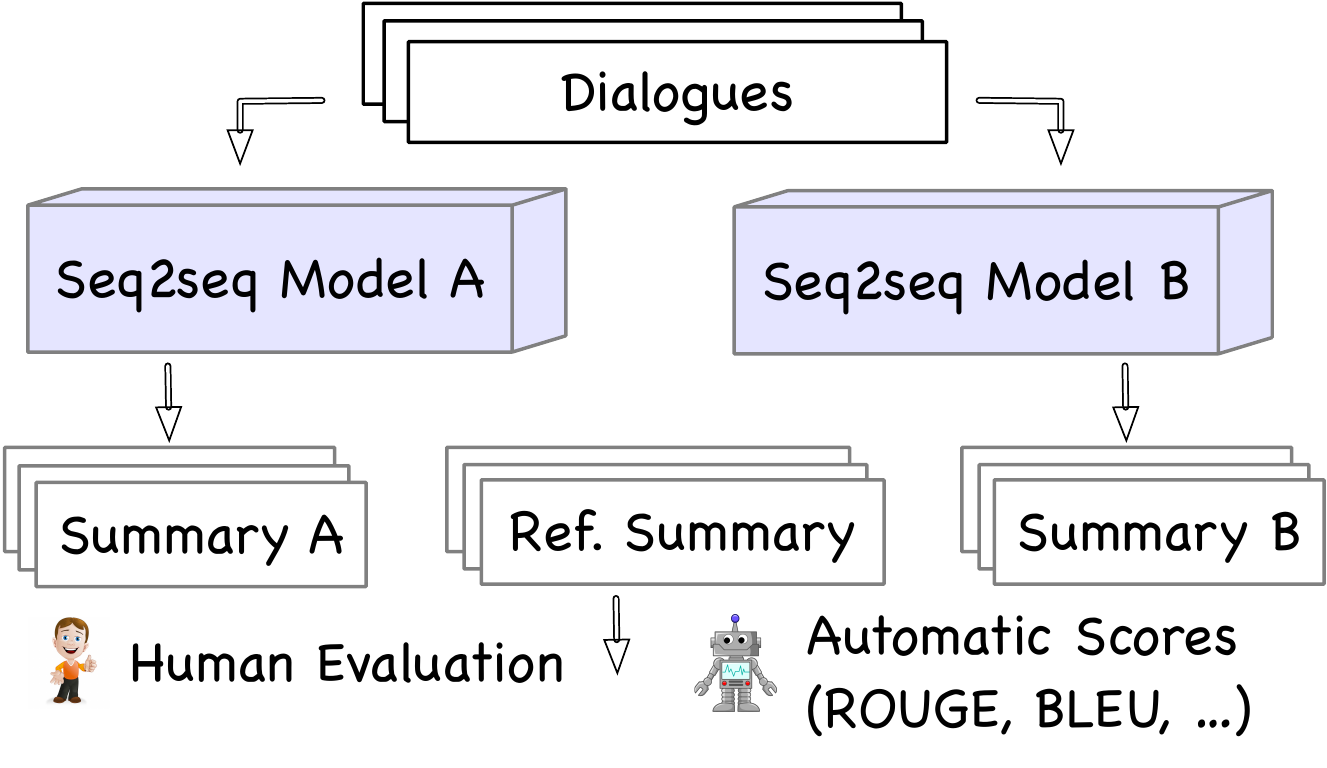}
     \caption{Sample-level evaluation schema.}
     \label{fig:sample-based-fact}
 \end{subfigure}
 \hfill
 \begin{subfigure}[b]{0.47\textwidth}
     \centering
     \includegraphics[width=\textwidth]{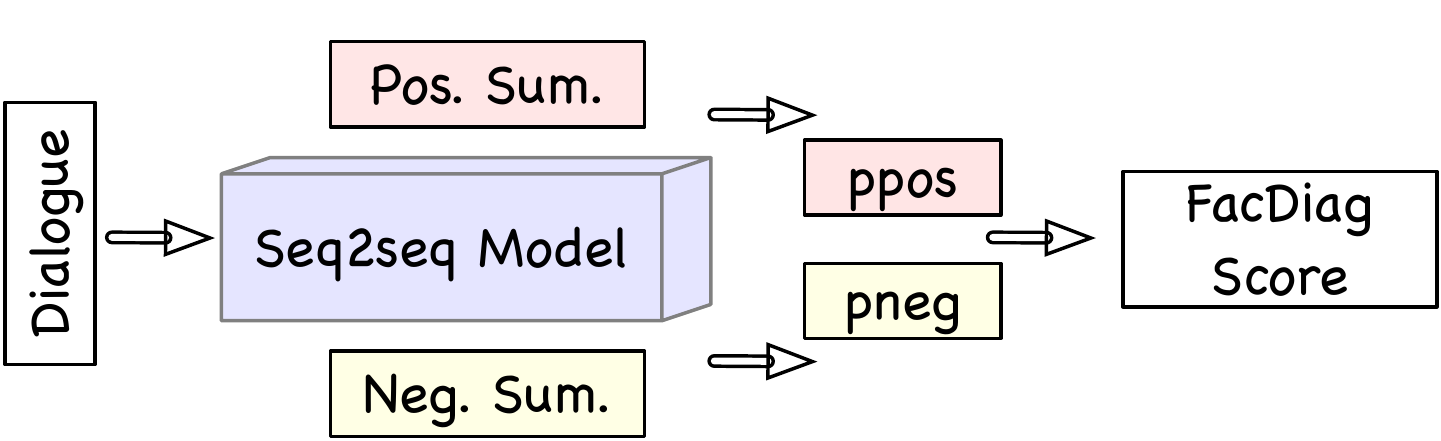}
     \caption{Model-level evaluation schema.}
     \label{fig:model-based-fact}
 \end{subfigure}
    \caption{An illustration of two types of evaluation paradigms. 
    }
    \label{fig:evaluation-types}
\end{figure}

\section{Model-level Faithfulness Evaluation}
\label{sec:faceval}

    Some efforts have been made toward sample-level factual error evaluation. An example is shown in Figure~\ref{fig:evaluation-types}. The sample-level evaluation methods are model-agnostic and examine a model solely based on its output sequences. Most existing evaluation methods, including ROUGE score, human evaluation and recent factual evaluation methods, belong to this type. One ultimate goal for factuality evaluation is to discriminate better summarization models. We propose directly probing models' generation probability with a constrained search space. First, FacEval generates a set of positive and negative samples with variant factual errors by rule-based transformations. Then, the generation probabilities of positive and negative summaries are compared for each dialogue. A better summarization model should be more likely to generate positive summaries than negative ones.


    
    
    \subsection{Dialogue-summary Pair Generation}
    \label{sec:dsp}
    
        We design transformations to synthesize negative samples with factual errors. Given the source and target text, one or more modifications are performed to the target text while referring to the information of the source text. It is because the frequently occurred errors are conceptual confusions from the source. Our designed transformations are listed as follows:
        
        \begin{itemize}
            \item \textbf{Speaker Swap (SS)}: We first spot the name of speakers from the source text by colon symbol and then swap the names at the target text.
            \item \textbf{Entity / Pronoun / Date / Number Swap (ES / PS / DS / NS)}: An NER system is first applied to both source and target text. The entities from the target text are randomly swapped with entities from the source text if they share the same entity type.
            \item \textbf{Negation (NG)}: Negation is performed using a set of hand-crafted rules. Auxiliary verbs are first scanned. Then, positive verbs are negated by adding not or n't. Similarly, negative sentences are inverted by negation removal.
        \end{itemize}
        
        First, we paraphrase the summary to create more positive samples through back-translation (\textbf{BT}). The Google Cloud API is leveraged for this task\footnote{\url{https://cloud.google.com/translate/}}. Then, we generate new summaries with factual errors by corrupting positive summaries, which means the summaries are treated as the target text, and the dialogue is the source text.
        
        Ideally, the negative summaries should be prone to errors generated in real-world scenarios. Therefore, our designed transformations try to mimic that. In the context of the analysis presented in Section~\ref{sec:analysis}, we have the following list of correspondences: 1) SS-SubObjE; 2) PS-ProE; 3) NG-NegE; 4) ES/DS/NS-ParE.
        
        


    \begin{figure}[t]
     \centering
        \includegraphics[width=0.45\textwidth]{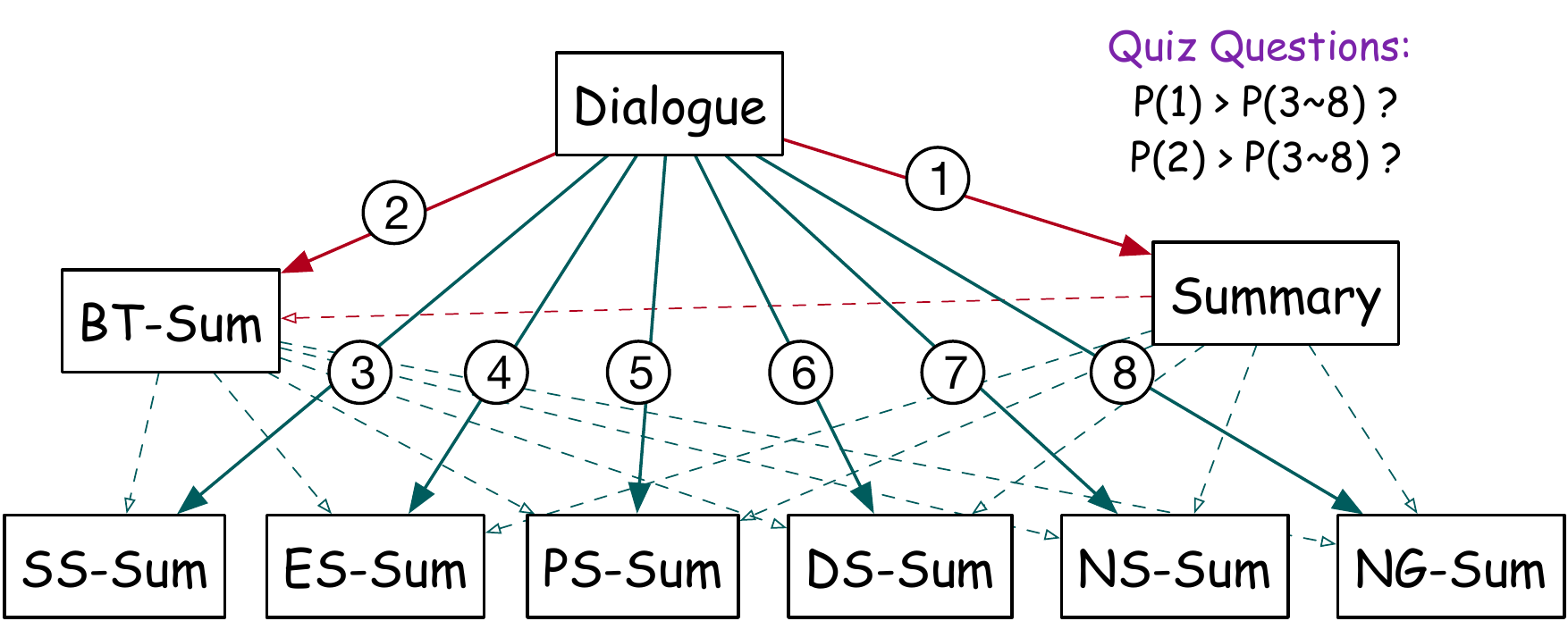}
        \caption{An illustration of comparing the generation probability of positive and negative samples. Solid and dashed lines refer to probability comparison and sample construction, respectively.}
        \label{fig:comp}
    \end{figure}

    \subsection{Comparison of Generation Probabilities}
        
        An illustration of probability comparison is shown in Figure~\ref{fig:comp}. Given a dialogue $D$, a summary $S=[y_1,...,y_L]$ and a summarization model $f_s(\cdot)$, we can compute a generation score ($GS$) for $D$-$S$ pair from the generation probability:
        \begin{align*}
        GS(S|D) & = \frac{1}{L^{\alpha}}log\ P(y_1,...,y_L|D) \\
                & = \frac{1}{L^{\alpha}} \sum_{t=1}^{L} log\ P(y_t|y_1,...,y_{t-1},D)
        \end{align*}
        where the generation probability for each token is as follows:
            \begin{align*}
            P(y_t|y_1,...,y_{t-1},D) & \\
                  = P(f_s&(y_1,...,y_{t-1},D)=y_t)
            \end{align*}
        We leverage the above generation score from decision process of beam search algorithm \cite{graves2012sequence}, where the sequence length is taken into consideration. In default, we set the length penalty parameter $\alpha$ as 1.0. 
        
        For dialogue $D_i$, there is positive summary set $S=[S_1,...,S_M]$ and negative summary set $\hat{S}=[\hat{S}_1,...,\hat{S}_N]$. We evaluate the number of times the positive samples have higher scores than the negative samples concerning the same dialogue. The factuality score ($FS$) of model $f_s(\cdot)$ is then computed as follows:
        \begin{align*}
            FS(f_s) & = \frac{1}{|D|}\sum_{i=1}^{|D|} \frac{1}{MN} \cdot  \\    &\sum_{m=1}^{M}\sum_{n=1}^{N}\mathbbm{1}[GS(S_m|D_i)>GS(\hat{S}_n|D_i)]
        \end{align*}
        where $|D|$ is the number of dialogues. 
            
  
    \subsection{Evaluation Preparation}
    \label{sec:ep}
    
        A series of models need to be prepared with different faithfulness capabilities to evaluate the effectiveness of model-level evaluation methods. One option is to collect as many well-trained models as possible and refer to human annotations to rank models based on factuality. However, it is hard to reach a high agreement and may not be trustworthy with limited annotators, as indicated by \citet{falke2019ranking} and \citet{kryscinski2019evaluating}. Therefore, instead, we construct a series of models using the following two ad-hoc methods:
        
            \noindent 
            \textbf{Limited data training (LDT).} One joint agreement is that more training data lead to better model performance. Therefore, we train 20 models using different proportions of the training data from 5\% to 100\%.
            
            \noindent 
            \textbf{Mixed data training (MDT).} In this setting, we randomly replace the human-labelled training samples with noisy ones. The noisy samples are created by corrupting only the dialogue using transformations introduced in Section~\ref{sec:dsp}. Here, the source and target are both the dialogue. The trained model is more likely to be confused and generate more factual errors with noisy data. Here, we obtain 21 models with different replacement ratios from 100\% to 0\%.
            
            
    
        LDT will cause a model to be less competent for generation in all aspects. In comparison, MDT will lead the model to generate summaries with more factuality errors while less affecting other properties like fluency. Therefore, we expect a better factuality evaluator to correlate more with MDT models. All correlations are computed on model-level instead of sample-level judgements. 
        

    
    \begin{table}[t]
      \centering
      \begin{adjustbox}{width=0.46\textwidth,center}
        \begin{tabular}{ c | c | c | c | c }
            \toprule
            & \emph{\# Diag} & \emph{\# Spk} & \emph{\# Turn}& \emph{Sum. Len.} \\ \midrule
            Train & 14,732 & 2.40 & 11.17 & 23.4 \\
            Val & 818  & 2.39 & 10.83 & 23.4 \\
            Test & 819 & 2.36 & 11.25 & 23.1 \\\bottomrule
        \end{tabular}
      \end{adjustbox}
      \caption{The detailed statistics of the SAMSum dataset. The header refers to the number of dialogues, the average number of speakers, the average number of dialogue turns, and the average summary lengths.}
      \label{tab:samsum}
    \end{table}

\section{Experiments}

        

    \begin{table*}[t]
        \centering
        \begin{adjustbox}{width=0.92\textwidth,center}
        \begin{tabular}{ c | c | c c c c c c | c }
        \toprule
         \textbf{Train. Strategy} & \textbf{Model}  & \textbf{NG} & \textbf{PS} & \textbf{SS} & \textbf{ES} & \textbf{DS} & \textbf{NS}  & \textbf{\emph{All}} \\ \midrule
         \multirow{4}{*}{\emph{LDT}}    & \emph{BART\textsubscript{Large}}  & 79.79 & 31.28  & 87.82 & -18.65 & 77.89  & -2.41  & 84.70 \\
                                        & \emph{BART\textsubscript{Base}}   & 89.62 & 75.64 & 98.65 & 73.83  & 40.84  & -2.26  & 99.40 \\ \cline{2-9}
                                        & \emph{T5\textsubscript{Base}}     & 63.16 & 46.62 & 93.98 & 50.68  & 73.38  & 7.98   & 89.62 \\
                                        & \emph{T5\textsubscript{Small}}    & 96.54 & 90.53 & 97.29 & 91.13  & 87.52  & 43.98  & 98.05 \\ \midrule
         \multirow{4}{*}{\emph{MDT}}    & \emph{BART\textsubscript{Large}}  & 64.03 & 53.12 & 99.87 & 67.14  & -21.30 & 35.84  & 99.74 \\
                                        & \emph{BART\textsubscript{Base}}   & 40.91 & 95.97 & 100.0 & 77.53  & 0.78   & 22.18  & 100.0 \\ \cline{2-9}
                                        & \emph{T5\textsubscript{Base}}     & 79.22 & 90.91 & 99.87 & 92.73  & -4.29  & 43.34  & 99.87 \\
                                        & \emph{T5\textsubscript{Small}}    & 89.74 & 88.31 & 100.0 & 90.78  & -12.27 & -38.26 & 99.74  \\ \midrule
           \midrule
        \multicolumn{2}{c|}{\# Neg Samples} & 3,094 & 2,990 & 2,100 & 643 & 547 & 98 & 9,472 \\\bottomrule
        \end{tabular}
        \end{adjustbox}
        \caption{Detailed correlation analysis between model series and negative sample types. For each column, one negative type is involved. `\emph{all}' indicates the usage of all negative types.}
        \label{tab:detail_fac}
    \end{table*}
    
    \subsection{Experimental Settings}
        
        SAMSum dataset \cite{gliwa2019samsum} is used for all experiments. It consists of 16,369 dialogue-summary pairs written by expert linguistics. One human-written reference summary is provided for each dialogue. The detailed dataset statistics are listed in Table~\ref{tab:samsum}. The samples from the test set are used for all evaluation methods.
        
        For backbone models, we exam with \emph{BART\textsubscript{Large}}, \emph{BART\textsubscript{Base}}\cite{lewis2019bart}, \emph{T5\textsubscript{Base}} and \emph{T5\textsubscript{Small}} \cite{raffel2019exploring}, which are SOTA summarization models. Each model is trained with both LDT and MDT methods. As a result, we obtained 164 trained models, divided into eight groups. 
        The models in each group are associated with increasing levels of capabilities. 
        The Spearman's rank correlation coefficient ($\rho$) between these models and evaluation scores is reported. For sample-based evaluation methods, the scores on all test set samples are averaged as the model-level performance. We ensure all models are appropriately trained and avoid training collapses by examining their ROUGE scores. The best hyper-parameters are used and kept the same for models from the same group.
        

    \subsection{Results and Analysis}
        
        Table~\ref{tab:detail_fac} shows the fine-grained results of FacEval. 
        First, we found that FacEval has a higher correlation with MDT models than LDT models. The LDT models are less competent in all aspects as fewer data are involved with training. The generated summaries are weaker in multiple elements, including factuality, fluency, coherence, and granularity. In contrast, the MDT models mainly deteriorate in factuality with factually corrupted training data. Therefore, it is desired that FacEval shows a higher correlation to MDT models.
        
        Second, when considering each negative sample type, a relatively higher correlation is shown with negation (NG), pronoun swap (PS) and speaker swap (SS). It is because more comparison pairs are created with these methods. Also, for chit-chat dialogues, almost all summaries contain reasoning concerning speakers and personnel in the dialogue. And the confirmation of action is happening in multiple utterances. As a result, these several error types are more commonly witnessed in dialogue summarization, as illustrated in Figure~\ref{fig:error_ratio}.
        In contrast, the negative pairs generated by entity swap (ES), date swap (DS) and number swap (NS) show a lower correlation. It is because these samples are more related to particular errors which appear in various formats and are more challenging to simulate. Even though solely considering these samples shows a lower correlation, we still include them in the overall comparison process to have a more comprehensive evaluation. 

    \begin{table*}[t]
      \centering
      \begin{adjustbox}{width=1.00\textwidth,center}
        \begin{tabular}{ c | c  c  c  c | c  c  c  c | c }
            \toprule 
            \textbf{Train. Strategy} & \multicolumn{4}{c|}{LDT} & \multicolumn{4}{c|}{MDT} & \multirow{2}{*}{\textbf{\emph{Avg.}}} \\\cline{1-9}
            \textbf{Model} & \emph{BART\textsubscript{Large}} & \emph{BART\textsubscript{Base}} & \emph{T5\textsubscript{Base}} & \emph{T5\textsubscript{Small}} & \emph{BART\textsubscript{Large}} & \emph{BART\textsubscript{Base}} & \emph{T5\textsubscript{Base}} & \emph{T5\textsubscript{Small}} \\ \midrule
            \multicolumn{10}{c}{\textit{Non-Factual Evaluation Schema}} \\ \midrule
            ROUGE-1     & 81.35 & 95.79 & 94.44 & 95.94 & 84.16 & 91.04 & 95.58 & 85.84 & 90.52 \\
            ROUGE-2     & 86.77 & 96.84 & 96.39 & 96.09 & 90.13 & 97.01 & 95.45 & 93.90 & 94.07 \\
            ROUGE-L     & 75.64 & 96.24 & 96.39 & 92.63 & 86.23 & 98.31 & 97.14 & 94.16 & 92.09 \\
            BLEU        & \textbf{91.88} & 90.08 & 92.33 & 86.02 & 89.87 & 94.16 & 93.38 & 84.03 & 90.22 \\
            BERTScore   & 88.87 & 97.14 & \textbf{97.29} & 95.79 & 91.69 & 94.42 & 95.97 & 92.47 & 94.20 \\ \midrule
            \multicolumn{10}{c}{\textit{Factual Evaluation Schema}} \\\midrule
            FactCC\textsubscript{v1} & --- & --- & --- & --- & --- & --- & --- & --- & --- \\
            FactCC\textsubscript{v2} & 82.39 & 84.57 & 42.45 & 97.07 & 96.01 & 99.22 & 98.27 & \textbf{100.0} & 87.50 \\
            FEQA        & 6.02 & 30.08 & -60.15 & 33.23 & 57.92 & 54.29 & 75.06 & 85.58 & 35.25 \\
            NLI         & 39.40 & 31.28 & 93.08 & 90.53 & 91.17 & 82.99 & 93.77 & 92.99 & 76.90 \\ \midrule \midrule
            \emph{FacEval} (ours) & 83.70 & \textbf{99.40} & 89.62 & \textbf{98.05} & \textbf{99.74} & \textbf{100.0} & \textbf{99.87} & 99.74 & \textbf{96.27} \\
            \bottomrule
        \end{tabular}
      \end{adjustbox}
      \caption{Comparison of a series of automatic evaluation metrics. The result shown is Spearman's rank correlation between model ranks and predicted scores.}
      \label{tab:automatic_eval}
    \end{table*}

    \subsection{Comparison with Other Metrics}
    
        We include a list of popular evaluation methods for summarization to compare our evaluation schema with existing ones. It contains three generic evaluation methods and four dedicated faithfulness evaluation methods.
    
        \subsubsection{Baseline Metrics}
            
            Three generic evaluation methods are as follows:
            
            
            \textbf{ROUGE}~\cite{lin2004rouge} score is the default evaluation metric for summarization. We experiment with the F-measure of ROUGE-1, ROUGE-2 and ROUGE-L\footnote{The py-rouge package is used. Different implementations can lead to discrepant results, as discussed in~\citet{feng2021survey}. \url{pypi.org/project/py-rouge/}}, which are derived from the uni-gram overlap, the bi-grams overlap and the longest common subsequence (LCS) between generated and reference summaries, respectively.
                
            \textbf{BLEU}~\cite{papineni2002bleu} score is the primary evaluation metric for machine translation. It is mainly designed for corpus-level similarity computation derived from n-gram overlaps. In the following experiments, we report the most commonly used BLEU-4 score.
                
            \textbf{BERTScore}~\cite{zhang2019bertscore} leverages the pre-trained contextual embeddings from BERT and computes the similarity between text sequences by matching words in candidate and reference by cosine similarity.
                
            
            \noindent Four faithful evaluation methods are as follows:
            
            
            \textbf{FactCC}\textsubscript{v1} \cite{kryscinski2019evaluating} first augment summaries by applying rule-based transformations from the document sentences and fine-tune a pre-trained language model BERT to classify whether the summary is consistent or inconsistent with the documents. It is initially trained in the news summarization domain.
                
            \textbf{FactCC}\textsubscript{v2} is an adapted FactCC\textsubscript{v1} to the dialogue domain by us. The negative summaries are generated using our transformations discussed in Section~\ref{sec:dsp}. We train a \emph{T5\textsubscript{Small}} model as the classifier and take dialogue and summary as input to predict their consistency.
                
            \textbf{FEQA} \cite{durmus2020feqa} is a question generation and answering method for faithfulness evaluation. It first extracts question-answer pairs from summaries with pre-trained models. Then, a QA model pulls answers from the document with the same questions. A matching method is deployed to measure the similarity between both answer responses from the summary and document as the factuality score. Note that the model is designed for documents. 
                
            \textbf{NLI} \cite{falke2019ranking} is an entailment-based method which takes the maximal entailment probability between summary and document sentence as the factual consistency score. As no dialogue-based entailment model is available, we compute the entailment probability between reference and generated summaries with a BERT-based entailment model trained on SNLI and MultiNLI datasets. 
            
            
        \subsubsection{Results and Analysis}
            
            The experimental results are shown in Table~\ref{tab:automatic_eval}.
            
            \noindent\textbf{Non-factual Evaluator}: The non-factual evaluation methods measure the similarity between reference and generated summaries. ROUGE and BLEU are derived from n-gram overlaps, which indicate the overall generation quality. It is expected that evaluators have a reasonable correlation with LDT models as training with fewer data will resulting quality degradation of the summary in all aspects. For MDT models, they also show a good correlation. We observe that R-2 and R-L are better indicators than R-1 for factuality evaluation. It is because simply replacing isolated tokens can easily change the factual correctness of a summary without much influence on the R-1 score. 
            
            \noindent\textbf{Factual Evaluator}: As FactCC\textsubscript{v1} is trained for the news summarization, we found that the released model is incapable of making predictions for dialogues. Similarly, FEQA is not a good indicator of model performance because the question and answer generation models are optimized for documents, which limits its transferability to the dialogue domain. In comparison, FactCC\textsubscript{v2} and NLI are better evaluation methods for factuality and can make good predictions on MDT models.
           
            \begin{table*}[t]
              \centering
              \begin{adjustbox}{width=0.75\textwidth,center}
                \begin{tabular}{ l | c | c | c | c | c | c | c | c }
                    \toprule
                    \textbf{Model} & \textbf{\# Params} & \textbf{All} & \textbf{NG} & \textbf{PS} & \textbf{SS} & \textbf{ES} & \textbf{DS} & \textbf{NS} \\ \midrule
                    \emph{BART\textsubscript{Large}} & 400M & 86.87 & 89.03 & 89.08 & 84.19 & 89.46 & 84.22 & 68.37 \\
                    \emph{BART\textsubscript{Base}} & 140M & 85.73 & 87.73 & 89.43 & 79.46 & 89.96 & 82.50 & 68.63 \\ \midrule
                    \emph{T5\textsubscript{Base}} & 220M & 85.32 & 86.81 & 88.27 & 79.92 & 90.71 & 84.90 & 67.75 \\ 
                    \emph{T5\textsubscript{Small}} & 60M & 79.87 & 82.09 & 85.94 & 69.68 & 87.17 & 79.88 & 62.96 \\ \bottomrule
                \end{tabular}
              \end{adjustbox}
              \caption{Benchmarking results on 4 different models}
              \label{tab:benchmarking}
            \end{table*}
            
            \noindent\textbf{FacEval Properties}: FacEval is the only model-level evaluation schema. The examined model requires reasonable predictions on single sentences and differentiation between positive and negative pairs. Therefore, FacEval shows a strong correlation with LDT and MDT models. The exceptional performance on MDT models indicates that FacEval can effectively reflect model’s capability on factuality. 
            
        \subsubsection{Benchmarking and Future Directions}

            It is beneficial to provide benchmarking performance and analysis on popular dialogue summarization models. As discussed in Sec.~\ref{sec:analysis}, dedicated dialogue summarization models do not outperform their baseline models in terms of faithfulness. Therefore, we evaluate on \emph{T5} and \emph{BART} models instead. 
        
            The benchmarking results are shown in Tab~\ref{tab:benchmarking}. There are several interesting findings. First, \emph{BART\textsubscript{Large}} has the largest model size as well as the overall best performance. We can also conclude that larger pre-trained models are more faithful based on our evaluation. Second, \emph{BART} model is generally better than \emph{T5} in factuality with model size taken into consideration. This may be because that \emph{BART} is designed for the generation with various denoising objectives, while \emph{T5} is a sequence-to-sequence model for different tasks including but not limited to generation. Third, from fine-grained analysis, we can see that speaker information (from SS) is a major challenge for dialogue summarization. This is because dialogue involves multiple speakers and their roles are tightly involved in the ideal summarization. Therefore, how to improve the model's understanding capability on speaker roles is an interesting direction to explore~\cite{liu2021coreference}. Meantime, because some faithful errors are coming from lack of commonsense for existing models~\cite{wang2021inductive}. How to effectively combine hidden semantics~\cite{wang2020sbert} and well-structured knowledge~\cite{ge2022compounde} are also worth exploration.

\section{Conclusion}

    We believe our faithfulness analysis and evaluation method can facilitate the development of dialogue summarization systems. Instead of measuring faithfulness on generated summaries,  we directly assess the model's capability by multi-choice questions. We expect FacEval to be effectively extended to other generation scenarios.
    
\section{Limitations}

    The testing samples used in our method are obtained by rule-based transformations of the reference and back-translated summaries. It is still limited to the types of transformations designed. More transformation methods need to be proposed to have a comprehensive evaluation. To obtain more natural summaries, we can gather generated summaries and perform annotation by humans. The model can be evaluated in more aspects and closer to real-world scenarios with more available samples. 
    
    Verifying the effectiveness of the model-level evaluation schema requires various models and their corresponding rankings. However, such model rankings are currently unavailable because 1) there are not enough varieties of dialogue summarization models as it is still a developing field; 2) the annotations on the faithfulness of dialogue summaries are not adequate. Therefore, in this work, we refer to heuristic methods to manually create a series of models with desired capability levels. When new evaluators are proposed, the best practice is to leverage model-level human rankings for performance benchmarking.

\section*{Acknowledgement}

    This research is supported by the Agency for Science, Technology and Research (A*STAR) under its AME Programmatic Funding Scheme (Project No. A18A2b0046) and Science and Engineering Research Council, Agency of Science, Technology and Research (A*STAR), Singapore, through the National Robotics Program under Human-Robot Interaction Phase 1 (Grant No. 192 25 00054).This work is also supported by the Internal Project Fund from Shenzhen Research Institute of Big Data under Grant T00120220002.

\bibliography{anthology,custom}
\bibliographystyle{acl_natbib}

\end{document}